\documentclass[lettersize,journal]{IEEEtran}
\usepackage{amsmath,amsfonts}
\usepackage{algorithmic}
\usepackage{algorithm}
\usepackage{array}
\usepackage{textcomp}
\usepackage{stfloats}
\usepackage{url}
\usepackage{verbatim}
\usepackage{graphicx}
\usepackage{cite}
\usepackage{subcaption} % For creating subtables
\usepackage{booktabs}
\usepackage{tablefootnote}
\usepackage{threeparttable}
\usepackage{graphicx}
\usepackage{float}
\usepackage{booktabs} % for better table appearance
\usepackage{multirow} % for vertical centering of cells
\usepackage{xcolor}
\usepackage{varwidth}
\usepackage[export]{adjustbox}
\usepackage{orcidlink}
\begin{document}

\title{PandaSkill - Player Performance and Skill Rating in Esports: Application to League of Legends}


%%%%%%%%%%%%%%%%%%%%%%%%%%%%%%%%%%%%%%%%%%%%%%%%%%%%%%%%%
%%%%%%%%%%%%%%%%%%%%%%%%%%%%%%%%%%%%%%%%%%%%%%%%%%%%%%%%%
%%%%%%%%%%%%%%         TODO        %%%%%%%%%%%%%%%%%%%%%%
%%%%%%%%%%%%%%%%%%%%%%%%%%%%%%%%%%%%%%%%%%%%%%%%%%%%%%%%%
%%%%%%%%%%%%%%%%%%%%%%%%%%%%%%%%%%%%%%%%%%%%%%%%%%%%%%%%%
% - make sure data in app is the same as in paper before doing final submission.

\author{Maxime~De~Bois \orcidlink{0000-0002-4181-2422},
         Flora Parmentier, Raphaël Puget, Matthew Tanti, and Jordan Peltier
\thanks{All authors are from PandaScore. Email: maxime.debois@pandascore.co}
\thanks{This work has been submitted to the IEEE for possible publication. Copyright may be transferred without notice, after which this version may no longer be accessible.}
}

%%%%%%%%%%%%%%%%%%%%%%%%%%%%%%%%%%%%%%%%%%%%%%%%%%%%%%%%%
%%%%%%%%%%%%%%%%%%%%%%%%%%%%%%%%%%%%%%%%%%%%%%%%%%%%%%%%%
%%%%%%%%%%%%%%         TODO       %%%%%%%%%%%%%%%%%%%%%%%
%%%%%%%%%%%%%%%%%%%%%%%%%%%%%%%%%%%%%%%%%%%%%%%%%%%%%%%%%
%%%%%%%%%%%%%%%%%%%%%%%%%%%%%%%%%%%%%%%%%%%%%%%%%%%%%%%%%

        % <-this % stops a space
% \thanks{TBW}% <-this % stops a space

% \author{\textit{Anonymous}}

\markboth{Journal of \LaTeX\ Class Files,~Vol.~14, No.~8, August~2021}%
{Shell \MakeLowercase{\textit{et al.}}: A Sample Article Using IEEEtran.cls for IEEE Journals}


\maketitle

\begin{abstract}
To take the esports scene to the next level, we introduce PandaSkill, a framework for assessing player performance and skill rating. Traditional rating systems like Elo and TrueSkill often overlook individual contributions and face challenges in professional esports due to limited game data and fragmented competitive scenes. PandaSkill leverages machine learning to estimate in-game player performance from individual player statistics. Each in-game role is modeled independently, ensuring a fair comparison between them. Then, using these performance scores, PandaSkill updates the player skill ratings using the Bayesian framework OpenSkill in a free-for-all setting. In this setting, skill ratings are updated solely based on performance scores rather than game outcomes, hightlighting individual contributions. To address the challenge of isolated rating pools that hinder cross-regional comparisons, PandaSkill introduces a dual-rating system that combines players' regional ratings with a meta-rating representing each region's overall skill level. Applying PandaSkill to five years of professional League of Legends matches worldwide, we show that our method produces skill ratings that better predict game outcomes and align more closely with expert opinions compared to existing methods.
\end{abstract}

\begin{IEEEkeywords}
Esports, League of Legends, player performance, skill rating, TrueSkill, OpenSkill.
\end{IEEEkeywords}

\section{Introduction}

\IEEEPARstart{R}{ating} and ranking athletes has always been an integral part of competitive sports \cite{hvattum2019comprehensive, lasek2013predictive}. Providing simple and quantifiable measures of player performance is beneficial for both professionals and fans. Such systems can be used by the industry for coaching and scouting future talents or to drive fan engagement by unveiling key game insights \cite{bornn2018soccer}.

Recently, the increased availability of data has opened the way to more advanced, data-driven systems. Unlike traditional rating systems that rely on predefined formulas \cite{thorn2015hidden}, data-driven approaches leverage large and fine-grained datasets with machine learning techniques to analyze player performance in a game \cite{brooks2016developing, pappalardo2019playerank, rein2016big, decroos2019actions}. These methods can model more complex aspects of the game, such as the role of a given player or their contribution to the outcome of the game \cite{pappalardo2019playerank}.

A similar need for performance evaluation exists in esports, %such as League of Legends (LoL)
with titles such as League of Legends (LoL) or Counter-Strike, which require precise metrics to assess player skill \cite{varga2024esports}. Such systems can benefit players through better matchmaking \cite{delalleau2012beyond} and novel insights \cite{maymin2021smart, kleinman2021using}, professional teams with quantified assessments of team performance \cite{sabtan2022current,sharpe2022indexing}, and other actors in the ecosystem such as betting operators \cite{xenopoulos2020valuing,pradhan2020power}.

%Compared to traditional sports, esports present unique opportunities but also challenges. On one hand, 
As video games, esports titles often have easier access to very detailed data compared to traditional sports, which often rely on tracking devices \cite{gudmundsson2017spatio} or human annotations \cite{pappalardo2019public}. In fact, publishers often provide means to access the data, such as demo files for Counter Strike or APIs for LoL and Dota 2 \cite{varga2024esports, kleinman2021using, xenopoulos2020valuing, pradhan2020power, demediuk2021performance}.

On the other hand, the game state is generally far more complex \cite{varga2024esports}. For example, in LoL, a player chooses between more than 150 characters (called \textit{Champions}) to play in a game. Each Champion has its own style of play and contributes differently to the team's victory. Players often play a specific role in a team, and even for a given role, it is common to see different playstyles develop \cite{demediuk2021performance}. Moreover, the game's balance is updated frequently, often adding new game mechanics. Something that was effective in one version might not be in the next.

%On the other hand, the game state is generally far more complex and constantly evolving \cite{varga2024esports}. For instance, in LoL, a player chooses between more than 150 characters (called \textit{Champions}) to play in a game. Each Champion has its own style of play and contributes differently to the team's victory. Moreover, the game's balance is updated frequently, often adding new game mechanics. Something that was effective in one version might not be in the next. Another consequence of the complex game state is that there is no single way to play the game \cite{demediuk2021performance}. Players often play a specific role in a team, and even for a given role, it is common to see different playstyles develop. 

Recent research has explored methods for evaluating player performance in games \cite{demediuk2021performance,maymin2021smart}. Although analyzing a player's past performances can provide a reasonable estimate of their general skill level, it fails to capture the complete picture. A player’s performance in a game is strongly influenced by the strength of their opponents. Facing weaker opponents may lead to exceptional performance, which might not accurately reflect their overall skill level. 
%Recent research has delved into the evaluation of player performance in a game \cite{demediuk2021performance,maymin2021smart}, but these studies present limitations with regard to estimating the general skill of a player. Indeed, the performance of a player in a game depends heavily on the strength of the opponents. If a player faces weaker opponents, they may perform very well, which is not very representative of their general skill level.
Algorithms such as Elo, TrueSkill, or OpenSkill \cite{elo1978rating, herbrich2006trueskill, Joshy2024} are usually used to estimate the skill of a player, based on the outcomes of past games and the strength of the opponents. However, these algorithms do not take into account the individual contribution of a player to the outcome of the game. Moreover, in esports, the lower number of games played per player, compared to non-professional gaming, slows down the calibration of these algorithms. Additionally, the scene is often fragmented, with some players rarely facing each other or even never at all. This raises the question of how to compare players from groups of players that have never played against each other. For instance, in LoL, apart from the two yearly international tournaments, players only face other players from the same region. This challenge, known as \textit{isolated rating pools}, where player skill ratings evolve independently within isolated regions, has been acknowledged in chess literature and remains an open problem \cite{glickman1999rating}.

To address these challenges, we propose \textit{PandaSkill}, a two-step framework to evaluate player performance and skill rating in esports. While the player performance represents the individual contribution of the player to the outcome of a given game, their skill rating represent how good they are in general compared to other players. We can summarize our contributions as follows:
\begin{itemize}
    \item We propose an estimation of the player's performance in a game by using machine learning models that predict the outcome of the game from the player's statistics. By directly linking the performance score to the predicted probability, we %enable a flexible evaluation of the player's performance. 
    make the methodology independent of the underlying model.
    Moreover, we ensure the fairness of the model by accounting for the different roles existing in the game.
    \item The skill ratings of the players are computed from all the past performance scores. We use the Bayesian framework OpenSkill in a novel free-for-all approach, where players are evaluated based on their performance relative to each other, independent of their teams. This way, the method focuses more on the players' individual performances.
    \item We highlight the challenge of isolated rating pools in esports, hurting cross-regional comparisons. To alleviate the issue, we propose that the player skill rating be a combination of their contextual rating (within their region) and a meta rating (representing the skill level of the region). Improving cross-regional comparisons, this dual-rating system enables a more accurate evaluation of players on a global scale.
    \item We apply the framework to five years of professional League of Legends matches from all the regions in the world. Compared to other methods in the literature, we demonstrate that our approach produces ratings that better predict the outcome of a game and are more in concordance with human experts.
    \item The source code and data associated with this paper can be found in the \href{https://github.com/PandaScore/PandaSkill}{PandaSkill GitHub repository}, alongside a \href{https://pandaskill.streamlit.app/}{web application} to visualize the player performances and skill ratings. By making our tools publicly available, we hope to encourage further research and practical applications in esports.
\end{itemize}

The remainder of the paper is structured as follows. In Section \ref{sec:relatedwork}, we review related work on player performance evaluation and rating systems in esports. We provide the reader with some background knowledge about League of Legends in Section \ref{sec:lol}. Section \ref{sec:methods} details our proposed methodology. In Section \ref{sec:experiments}, we present the experimental setup, data used, and results of our experiments. Finally, we discuss the implications of our findings and potential directions for future research in Section \ref{sec:conclusion}.

\section{Related Work}

\label{sec:relatedwork}

In this section, we review existing data-driven player performance and skill rating evaluation systems. We focus mostly on esports or video game applications. However, since the field is relatively new, when relevant, we also discuss work from traditional sports.

\subsection{Player Performance in a Game}

Evaluating player performance in esports involves a variety of methodologies, ranging from basic metrics to the use of machine learning models.

The easiest way is to directly use one player statistic computed at the end of the game, such as the kill-death-assist ratio (KDA) in LoL \cite{sapienza2018individual}, or kill-death ratio in Counter-Strike \cite{hltvrating}.

While easy to use, these metrics lack depth and do not tell the full story of what happened during the game. This led the community to craft more descriptive metrics, probably the most iconic of them in esports being the HLTV Rating for Counter-Strike \cite{varga2024esports, xenopoulos2020valuing}. First introduced in 2010 and later improved with versions 2.0 and now 2.1, it combines player end-game statistics into a single value representing how much better the player performed in a given game compared to the average \cite{hltvrating}.

Likewise, Demediuk \textit{et al.} proposed a framework to compute the current performance of a player during a game, which they called Performance Index (PI), for the esports game Dota 2 (which is similar to LoL) \cite{demediuk2021performance}. The PI is computed as the weighted sum of the percentiles of the player's statistics. The weights are derived from random forest models trained to predict the game outcome from the player statistics. Multiple random forest models were trained, one per role and predefined playstyle archetype to account for the complexity of the game.

A similar methodology has been followed in soccer with the PlayeRank framework \cite{pappalardo2019playerank}. They trained a single Support Vector Classifier (SVC) model to predict the game outcome from team statistics (e.g., number of passes). The performance of a player is then computed as the share of the team statistic that can be attributed to the player. They also explored a role-based version of their framework, where the weights are derived from players of similar roles.

Alternatively, researchers have explored the creation of more descriptive metrics, albeit not necessarily fully representative of the player performance. For instance, for Counter-Strike, Xenopoulos \textit{et al.} developed the Win Probability Added (WPA) framework to value players' actions based on changes in their team's chances of winning \cite{xenopoulos2020valuing}. In LoL, Maymin proposed identifying \textit{smart kills} and \textit{worthless deaths} from the game events, as these are considered to have a strong impact on the outcome of the game \cite{maymin2021smart}.

\subsection{Player Skill Rating}

The simplest way to rate the skill of players and rank them is to average player metrics representative of their past performance. For instance, PlayeRank used the exponentially weighted moving average (EWMA) technique over the performance scores of the player in past games, weighting recent performances higher \cite{pappalardo2019playerank}. On the other hand, Xenopoulos \textit{et al.} used averages of their proposed WPA metric and HLTV Rating 2.0 over a rolling window of one month \cite{xenopoulos2020valuing}.

However, averaging on recent player's performances does not take into account the strength of the opposing players in the calculation. Consequently, a high rating can be achieved by consistently competing against weaker opponents, resulting in an inflated ranking. In the context of chess, the Elo rating system has been widely adopted as a method to address this issue \cite{elo1978rating}. Elo adjusts a player's rating based on both the outcome of the match and the relative strength of the opponent, ensuring that victories against stronger opponents yield a greater rating increase, while wins against weaker opponents have a smaller impact.

The Glicko and later the Glicko-2 rating systems improve upon Elo by introducing a measure of uncertainty into a player's rating, making it one of the first Bayesian rating systems \cite{glickman1995glicko, glickman2012example}. The uncertainty represents the reliability of the rating. Each player starts with a high uncertainty value which slowly decreases game after game. Both Elo and Glicko ratings have been extensively used with applications in football \cite{hvattum2019comprehensive,lasek2013predictive}, in casual LoL matchmaking \cite{varga2024esports}, and in professional Dota 2 teams rankings \cite{pradhan2020power}.

Nonetheless, both Elo and Glicko are designed for two-player games, which most esports titles are not. When applied to two-team games (which is very common), the individuality of the players is not considered.
To address this problem in the context of casual matchmaking, Microsoft Research developed TrueSkill and TrueSkill2 to be used in first-person shooters (FPS) such as Halo or Gears of War \cite{herbrich2006trueskill, minka2018trueskill}. It differs from Elo and Glicko by updating the rating after each game instead of a given time period, and by working in multi-team/multi-player contexts. One of the additions of TrueSkill2 is the use of additional metrics about the players, such as the number of kills scored or the tendency to quit the game. 

A key downside of TrueSkill, however, is that it remains a proprietary framework. To provide a free and open-source alternative, OpenSkill was developed \cite{Joshy2024}. Thanks to the use of Bayesian approximation, OpenSkill is also faster, which is convenient in video games when there are a lot of matches to compute \cite{weng2011bayesian}.

Another alternative to TrueSkill has been proposed by Delalleau \textit{et al.} to promote fun in casual matchmaking for the FPS Ghost Recon Online \cite{delalleau2012beyond}. They used a neural network to predict both balance and player enjoyment from the players' profiles (previous game statistics like number of kills or firing accuracy). Pradhan \textit{et al.} explored a different approach, using a multi-criteria decision-making system to rank Dota 2 teams in a given season based on team metrics \cite{pradhan2020power}.

\subsection{Position of Our Work}

In this paper, we propose a two-step end-to-end player skill rating evaluation. First, we compute the performance score of each player in a game from their end-game statistics. Then, we update the skill ratings of the players based on their respective performance in the game. This methodology is applied to five years of professional League of Legends.

%Regarding the performance score calculation, all existing methods end up being a linear combination of the end-game statistics. Instead, by using a machine learning model that predicts the game outcome from the player's end-game statistics and by directly linking the player's performance score to the predicted probability, we enable a more complex estimation of the performance score.

When calculating performance scores, our main goal is to base the evaluation on features that are most strongly tied to a team’s chances of winning. Existing methods approach this by combining end-game statistics in a linear way. In contrast, we tie the performance score directly to the predicted probability of winning, estimated by a machine learning model that predicts the outcome of the game from the player’s end-game statistics. This approach allows us to capture more complex relationships in the data, leading to a more accurate estimation of performance.

As for the evaluation of the skill ratings, to the best of our knowledge, there exists no framework that simultaneously uses the player performance in-game whilst taking into account the strength of the players in the game. Moreover, all rating systems are prone to the isolated rating pools challenge, where skill ratings of players evolve independently due to a low number of games between certain players \cite{glickman1999rating}. As we highlight the impact of this issue on League of Legends professional ratings, we address it by combining a player’s rating within their region with the rating of the region itself.

\section{Background on League of Legends}
\label{sec:lol}

\begin{figure}[htbp]
    \centering
    \includegraphics[width=0.35\textwidth]{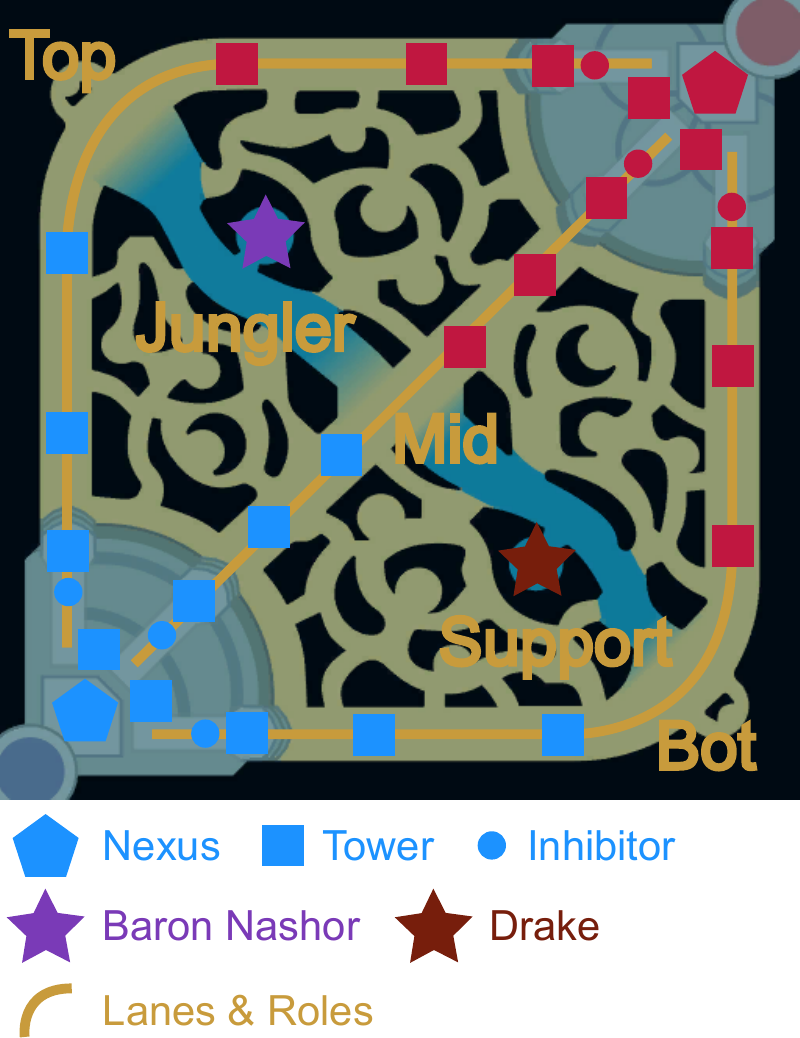}
    \caption{Annotated minimap of League of Legends.}
    \label{fig:summoners_rift}
\end{figure}

\subsection{Game Layout}

League of Legends (LoL) is a competitive Multiplayer Online Battle Arena (MOBA) game. In LoL, two teams of five players each compete on a map called \textit{Summoner's Rift}, with the goal of destroying the enemy \textit{Nexus}, located inside the enemy base and surrounded by \textit{Towers} and \textit{Inhibitors}. The two bases are connected by three paths, commonly called lanes: the top lane, the mid lane, and the bottom lane. Figure \ref{fig:summoners_rift} shows the simplified LoL map. At regular time intervals, \textit{Minions} spawn in the two bases and push down the lanes, attacking nearby enemy buildings. To destroy the enemy base, the players assist the Minions in their efforts. Between the lanes is the jungle, where neutral monsters can be defeated. The strongest ones, such as the \textit{Drake} or the \textit{Baron Nashor}, provide powerful bonuses to the team that defeats them.

\subsection{Players}
Each player embodies a character called a \textit{Champion}. All Champions have a distinct set of abilities, strengths, and weaknesses. In a game, a player is also assigned a role. There are five different roles, one for each player: \textit{Top}, \textit{Jungler}, \textit{Mid}, \textit{Bottom}, and \textit{Support}. The roles represent both the area of the map the player is supposed to play in during the early stages of the game (see Figure \ref{fig:summoners_rift}) and the job of the player within the team. Here is a basic general understanding of the roles:

\begin{itemize}
\item {\textbf{Top}: plays in the top lane; strong and independent; often provides tankiness and map control to the team.}
\item{\textbf{Jungle}: plays between the lanes; moves around the map to apply pressure on the different lanes and help securing objectives.}
\item{\textbf{Mid}: plays in the mid lane; %strongest player in the early to mid game; 
has the most agency out of the laners in the early to mid game due to their central position in the map; often provides burst damage to the team.}
\item{\textbf{Bot}: plays with the Support player in the early game in the bottom lane; provides sustained damage to the team; scales well into the late game by obtaining more gold.}
\item{\textbf{Support}: plays with the Bot player in the early game in the bottom lane; does not need a lot of gold to perform; helps the team by providing vision%, healing, and various buffs.}
 and utility (e.g., healing, buffs, control).}
\end{itemize}

\subsection{General Game Strategy}

A typical game of League of Legends lasts between 20 and 50 minutes. During this time, each player tries to become stronger by leveling up and buying powerful items with the gold obtained by killing Minions, neutral monsters, enemy Champions or buildings. Securing important neutral objectives, such as Drakes or the Rift Herald, also provides valuable advantages. When they are strong enough, they can attempt to break into the enemy base by sieging and fighting the enemy team together.

\subsection{The Esports Ecosystem}

The highest competition tier of League of Legends is managed by the game publisher Riot Games. The world is divided into geographic regions, each having its own league and tournament format. Twice a year, on the occasions of the Mid-Season Invitational (MSI) and the Worlds, the best teams from each region compete against each other. For years, the international scene has been dominated by Korea and China, followed by Europe and North America. As the esports scene matures, other regions (e.g., Asia-Pacific) are starting to emerge as serious contenders.

\section{PandaSkill}

\label{sec:methods}

\begin{table*}[ht]
\small
\begin{tabular}{cc}
        \begin{tabular}{lcccc}
        \toprule
        \textbf{Player} & \textbf{Team} & \textbf{Role} & \textbf{PScore} & \textbf{Skill Rating} \\
        \toprule
            Viper & HLE & Bot&82.52 & 92.68 (+0.50)\\
            Faker & T1 & Mid & 81.02 & 84.43 (+0.45)\\
            Gumayusi & T1 & Bot&74.20 & 88.93 (+0.29)\\
            Peanut & HLE & Jungle& 69.88 & 91.84 (+0.20)\\
            Zeka & HLE & Mid& 59.09 & 89.61 (+0.17)\\
            Keria & T1 & Support & 58.76 & 90.26 (+0.05)\\ 
            Delight & HLE & Support&52.95 & 90.84 (-0.11)\\
            Oner & T1 & Jungle & 43.11 & 89.15 (-0.20)\\
            Doran & HLE & Top & 31.97 & 86.10 (-0.30)\\
            Zeus & T1 & Top&28.29 & 86.73 (-0.81)\\
        \bottomrule
        \end{tabular}
    &
    \includegraphics[scale=0.42,valign=m]{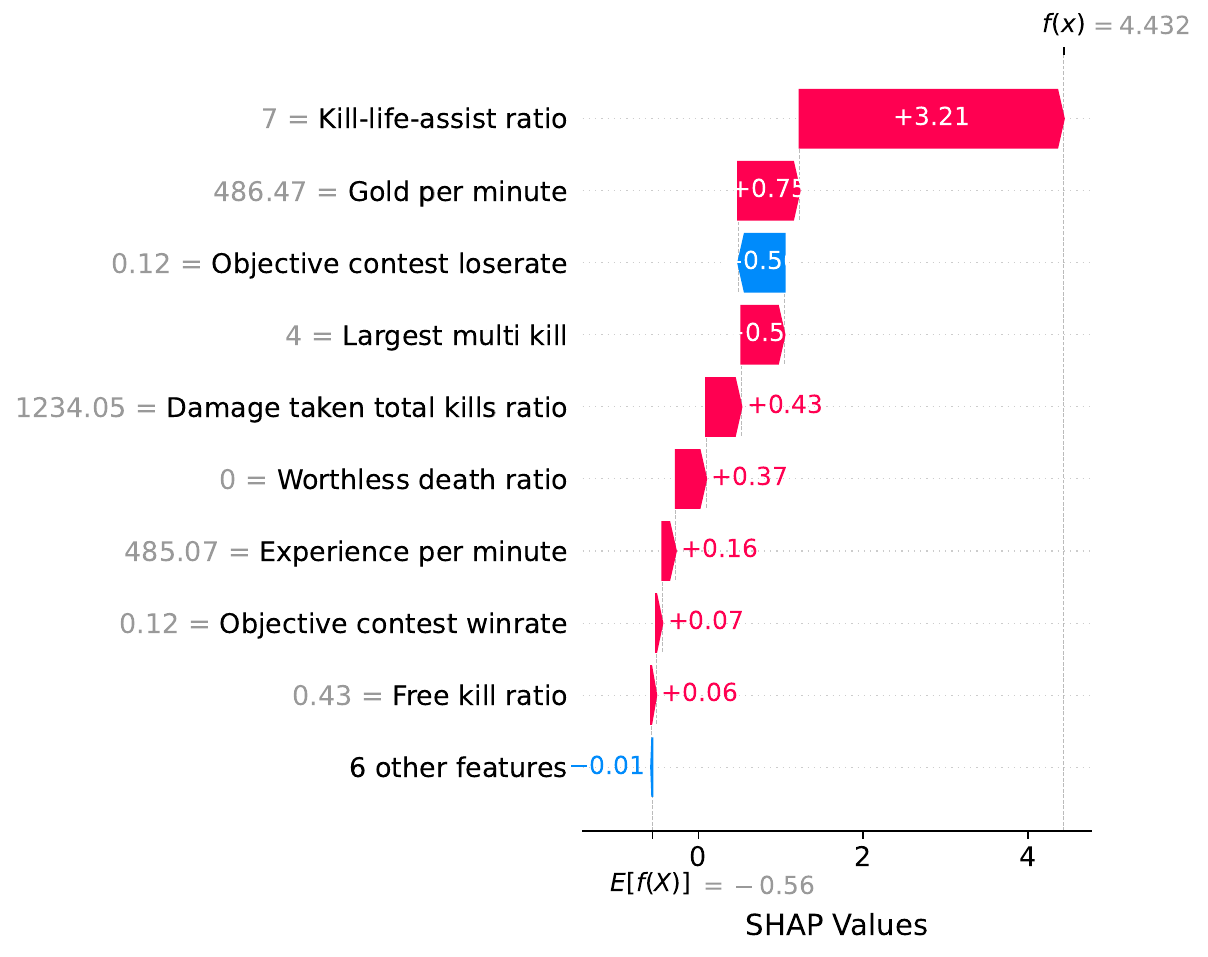} \\
    (a) Player performances (PScore) and updated skill ratings & (b) Explanation of Faker's PScore with SHAP values\\
    \end{tabular}
    \caption{Analysis of PandaSkill applied to game 2 of LCK Summer 2024 lower-bracket final, opposing Hanwha Life Esports (HLE) to T1 (victory of T1). The table highlights player skill ratings updated based on individual performance (PScore) rather than team outcomes.    
    }
\label{table:game_analysis}
\end{table*}

PandaSkill computes the players' skill ratings in two steps. First, after each game, it computes the performance of the players in the game. Then, it uses the player performances to update the players' skill ratings. Table \ref{table:game_analysis} shows a practical example of the framework applied to a game.

\subsection{Measuring the Performance Score of Players in a Game}

\label{sec:methods_perf}

Measuring a player's performance in a game is challenging due to the absence of a known ground truth. Similarly to the PlayeRank and PI frameworks, we use the outcome of the game as a proxy for the player's performance \cite{pappalardo2019playerank,demediuk2021performance}. The idea is that if a player played well during a game, their performance should be reflected both in their end-game statistics and in the outcome of the game.

More specifically, we propose the following methodology to calculate the Performance Score (PScore) of a player in a game:
\begin{enumerate}
    \item For each role, we train a machine learning model that predicts the probability of a player winning the game based on their end-game statistics (e.g., number of kills).
    \item The predicted probabilities are then transformed into percentiles. This transformation is learned for each model on the training data. %While the predicted probabilities can already be seen as a decent single-value representation of the player's performance, they depend heavily on the predictive capability of the model for that role. By transforming them into percentiles, the performance scores are uniformly distributed between 0 and 100. This transformation is learned for each model on the training data.    
\end{enumerate}

Compared to PI and PlayeRank, our approach has the following advantages:
\begin{itemize}
    \item Both PI and PlayeRank calculate the performance of the players by doing a linear combination of their end-game statistics, with weights that are derived from a model trained to predict the game outcome. By linking the performance scores directly to the model's predicted probabilities, we account for the model's non-linearity, enabling a more complex calculation of the player's performance score. %By linking the performance scores directly to the model's predicted probabilities, we enable a more complex calculation of the player's performance score.
    \item We ensure a fair cross-role comparison as all the scores are normalized on the same scale (from 0 to 100) by the percentile transformation, independent of the underlying model. This makes it possible to compare players with different roles, which is not the case for PlayeRank.
    \item Our approach is model-agnostic as long as the model can output probabilities. Both PI and PlayeRank work with specific models from which feature importance can be easily computed (i.e., random forest for PI and linear SVC for PlayeRank). %However, we note that, to get the best results, the model should have a good predictive capability, but also a good calibration. The latter is particularly important as the performance scores are directly derived from the probabilities.
\end{itemize}

Three qualities should be considered in order to choose which machine learning algorithm to use to compute the performance scores:
\begin{itemize}
    \item \textbf{Strong predicting capability}: A model without a good predicting capability will fail at capturing the nuances of the games.
    \item \textbf{Calibrated probabilities}: The model should produce probabilities that closely reflect reality; it is particularly important as performance scores are directly derived from the probabilities.
    \item \textbf{Interpretability}: It should be possible to explain how the performance scores are determined to ensure the system is transparent and trustworthy. 
\end{itemize}

In this paper, we used XGBoost models as they satisfy these qualities well.

\subsection{Estimating the Skill Rating of the Players}

\subsubsection{OpenSkill}

We base the evaluation of the skill rating of the players on the OpenSkill framework \cite{Joshy2024}. It has the advantage of being both open and working in multi-team or multi-player settings, which is not the case for Elo, Glicko, or TrueSkill \cite{elo1978rating, glickman1995glicko, herbrich2006trueskill}.

%Implementing the Plackett-Luce rating algorithm, OpenSkill is a Bayesian framework. The player's skill rating is modeled as a Gaussian distribution with the average skill of the player $\mu$ and an estimation of its uncertainty $\sigma$. Specifically, we represent the skill $S_i$ of player $i$ as:

OpenSkill models the player's skill rating as a Gaussian distribution with the average skill of the player $\mu$ and an estimation of its uncertainty $\sigma$. Specifically, we represent the skill $S_i$ of player $i$ as:

\begin{equation}
    S_i \sim \mathcal{N}(\mu_i, \sigma_i^2),
    \label{eq:player_skill}
\end{equation}

The ratings of all players are initialized to the default values of $\mu_i = 25$ and $\sigma_i = 25/3$, as it is generally done in Bayesian rating systems \cite{Joshy2024, herbrich2006trueskill}. After a game, both $\mu_i$ and $\sigma_i$ are updated based on the outcome of the player's team. The updates are derived from Bayesian inference, adjusting the player's skill distribution to reflect the new information provided by the game outcome. Formally, the update mechanism can be described by the following equation:

\begin{equation} 
(\boldsymbol{\mu}^{t+1}, \boldsymbol{\sigma}^{t+1}) = \Omega\left( \boldsymbol{\mu}^{t}, \boldsymbol{\sigma}^{t}, \text{game outcome} \right), 
\label{eq:openskill} 
\end{equation}

where:
\begin{itemize} 
\item ($\boldsymbol{\mu}^{t}$, $\boldsymbol{\sigma}^{t}$) represent the ratings of all the players in the game before the game at time step $t$;
\item ($\boldsymbol{\mu}^{t+1}$, $\boldsymbol{\sigma}^{t+1}$) denote their ratings after the game at time step $t+1$;
\item $\Omega$ is the Bayesian update made by the OpenSkill framework.
\end{itemize}

Since probabilistic estimates of skill ratings are often impractical to use directly (e.g., for creating rankings), they can be transformed into single-value estimates by taking the lower bound $\theta_i = \mu_i - 3 \cdot \sigma_i$. This provides a conservative estimate of the player's skill rating, ensuring a 99.7\% likelihood that the player's true rating is higher.

%The skill rating can be transformed into a single-value estimate by taking the lower bound $\theta_i = \mu_i - 3 \cdot \sigma_i$. It is a conservative estimate of the player's skill rating, representing a 99.7\% likelihood that the true rating of the player is higher.

\subsubsection{Free-For-All (FFA)}

\label{sec:ffa} 

Natively, OpenSkill does not use the performance of the players to adjust the rating updates (which is also the case for Elo, Glicko, or TrueSkill). To incorporate the evaluation of the performance of the players in the game into the rating updates, we suggest framing the ratings updates as a free-for-all (FFA) game. In this setting, all players compete against each other. After a game, they are ranked based on their individual PScore and their skill ratings are updated based on this ranking. The update mechanism can then be written as follows:

\begin{equation} 
(\boldsymbol{\mu}^{t+1}, \boldsymbol{\sigma}^{t+1}) = \Omega_{\text{FFA}}\left( \boldsymbol{\mu}^{t}, \boldsymbol{\sigma}^{t}, \textbf{PScore} \right), 
\label{eq:openskill_ffa} 
\end{equation}

where:
\begin{itemize}
    \item $\textbf{PScore}$ represents the performance scores of the players in the game;
    \item $\Omega_{\text{FFA}}$ is the Bayesian update made by the OpenSkill framework in the free-for-all setting.
\end{itemize}

 Using the FFA setting has several consequences:
\begin{itemize}
    \item The outcome of the game doesn't directly impact the update of the skill ratings. This means that a player can see their rating improve due to a good performance, even if they lost.
    \item Rating updates depend on the relative performance the player has achieved compared to the other players in the game. To increase their rating, the player needs to perform better than the other players (regardless of their team).
\end{itemize}

\subsubsection{Contextual and Meta Ratings}

\label{sec:meta_rating}

A significant challenge in global player ranking systems is the potential for ratings of players from different contexts (e.g., different regions or competition tiers) to evolve independently, creating isolated rating pools where direct comparisons become difficult. This is particularly problematic for some esports titles (e.g., League of Legends), where inter-regional play may be infrequent, limited to specific international tournaments.

To address this issue, we propose to model the skill rating of a player as the combination of two distinct ratings: a \emph{contextual rating} and a \emph{meta rating}. The contextual rating represents the player's skill within their context (e.g., region), while the meta rating represents the skill level of the context itself. This means that all players from the same context share the same meta rating.

We define the overall skill rating of a player as the sum of their contextual $(\mu_i^{\text{ctx}}, \sigma_i^{\text{ctx}})$ and meta $(\mu_i^{\text{meta}}, \sigma_i^{\text{meta}})$ ratings to capture both the skill of the player's within their context, and the context of the player:

\begin{equation}
    \mu_i = \mu_i^{\text{ctx}} + \mu_i^{\text{meta}},
    \label{eq:overall_mu}
\end{equation}

\begin{equation}
    \sigma_i^2 = {\sigma_i^{\text{ctx}}}^2 + {\sigma_i^{\text{meta}}}^2,
    \label{eq:overall_sigma}
\end{equation}

%where: \(\mu_i^{\text{ctx}}\) and \(\sigma_i^{\text{ctx}}\) are the contextual mean and standard deviation for player \(i\). and \(\mu_i^{\text{meta}}\) and \(\sigma_i^{\text{meta}}\) are the meta mean and standard deviation shared among all players in the same context.

After a game, using the OpenSkill framework, only one of the two ratings is updated depending on the context of the opponents in the match. If both teams share the same context, then the contextual rating of each player is updated. On the other hand, if the teams have different contexts (e.g., they are coming from different regions), we update the ratings of the teams' contexts. The following details how the updates are made in both cases:

\begin{itemize}
    \item \textbf{Updating contextual ratings}: If all opponents are from the same context, we update only the contextual ratings. Because all the players in the game share the same meta rating, we do not need to use the meta ratings to update the contextual ones. As a consequence, updating the contextual ratings can be seen as standard rating update as described earlier. Formally, we can write:

    \begin{equation} 
    (\boldsymbol{\mu}^{\text{ctx},t+1}, \boldsymbol{\sigma}^{\text{ctx},t+1}) = \Omega_{\text{FFA}}\left( \boldsymbol{\mu}^{\text{ctx},t}, \boldsymbol{\sigma}^{\text{ctx},t}, \textbf{PScore} \right), 
    \label{eq:openskill_ffa_ctx} 
    \end{equation}
    
    % Let the superscript $game$ denote de value used in the OpenSkill update for a given game. Here, the ratings used for the update are defined as:
    
    % \begin{equation}
    %     \mu_i^{\text{game}} = \mu_i^{\text{ctx}},
    %     \label{eq:ctx_mu_used}
    % \end{equation}

    % \begin{equation}
    %     \sigma_i^{\text{game}} = \sigma_i^{\text{ctx}},
    %     \label{eq:ctx_sigma_used}
    % \end{equation}
    
     %The ratings used for the update in the game are defined as:

    % \begin{equation}
    %     \mu_i^{\text{game}} = \mu_i^{\text{ctx}} + \theta_i^{\text{meta}},
    %     \label{eq:ctx_mu_used}
    % \end{equation}

    % \begin{equation}
    %     \sigma_i^{\text{game}} = \sigma_i^{\text{ctx}},
    %     \label{eq:ctx_sigma_used}
    % \end{equation}

    % where $\theta_i^{\text{meta}} = \mu_i^{\text{meta}} - 3 \cdot \sigma_i^{\text{meta}}$ is the lower bound of the context's skill rating.

    % The update of the contextual ratings is written as follows:

    % \begin{equation}
    %     \mu_i^{\text{ctx,t+1}} = \mu_i^{\text{game,t+1}},
    %     \label{eq:update_mu_ctx_game}
    % \end{equation}

    % \begin{equation}
    %     \sigma_i^{\text{ctx,t+1}} = \sigma_i^{\text{game,t+1}},
    %     \label{eq:update_sigma_ctx_game}
    % \end{equation}

    % In practice, because the same offset $\theta_i^{\text{meta}}$ is applied to all players in the game, it has no influence on the updated contextual ratings.
    
    \item \textbf{Updating meta ratings}: If the opponents are from different contexts, we update only the meta ratings. Contrary to the contextual rating updates, here we use the contextual ratings as offsets to the meta ratings to take into consideration the individual strength of the players in the game. The offsetting technique has been used in TrueSkill2 as an efficient way to balance the skill ratings of players playing as a squad and solo players \cite{minka2018trueskill}. The meta ratings update can be written as follows:

    \begin{equation} 
        \begin{split}
        (\boldsymbol{\mu}^{\text{meta},t+1}, \boldsymbol{\sigma}^{\text{meta},t+1}) = 
        \Omega_{\text{FFA}}\Big( & \boldsymbol{\mu}^{\text{meta},t} + \boldsymbol{\theta}^{\text{ctx},t}, \\
        & \boldsymbol{\sigma}^{\text{meta},t}, \textbf{PScore} \Big), 
        \end{split}
        \label{eq:openskill_ffa_meta} 
    \end{equation}

    % \begin{equation} 
    % (\boldsymbol{\mu}^{\text{meta},t+1}, \boldsymbol{\sigma}^{\text{meta},t+1}) = \Omega_{\text{FFA}}\left( \boldsymbol{\mu}^{\text{meta},t} + \boldsymbol{\theta}^{\text{ctx},t}, \boldsymbol{\sigma}^{\text{meta},t}, \textbf{PScore} \right), 
    % \label{eq:openskill_ffa_meta} 
    % \end{equation}

    where $\boldsymbol{\theta}^{\text{ctx},t}$ are the players lower-bound contextual ratings in the game.
    
    % The ratings used for the update in the game are defined as:

    % \begin{equation}
    %     \mu_i^{\text{game}} = \mu_i^{\text{meta}} + \theta_i^{\text{ctx}},
    %     \label{eq:meta_mu_used}
    % \end{equation}

    % \begin{equation}
    %     \sigma_i^{\text{game}} = \sigma_i^{\text{meta}},
    %     \label{eq:meta_sigma_used}
    % \end{equation}

    % where $\theta_i^{\text{ctx}} = \mu_i^{\text{ctx}} - 3 \cdot \sigma_i^{\text{ctx}}$ is the lower bound of the player's contextual skill rating estimate. It functions as an offset to $\mu_i^{\text{meta}}$ without modifying the variance of the Gaussian estimate. 

    However, there are two issues with Equation \ref{eq:openskill_ffa_meta}. First, all the resulting $\mu_i^{\text{meta}, t+1}$ are shifted by $\theta_i^{\text{ctx},t}$. We need to reverse the offset to preserve the original scale of $\mu_i^{\text{meta}, t}$. Moreover, $\boldsymbol{\mu}^{\text{meta}, t+1}$ contains multiple different meta rating updates for players sharing the same context. These variations arise from the variability in the players' contextual ratings and their individual performances in the game. In order to have the same updated meta rating for players that share the same context, we need to derive a single updated meta rating for each context in the game. We chose to average the updates for the same context as it ensures that the meta rating reflects the collective contributions and performances of all players associated with the context. The post-processing of the updates can be written as follows:
    
    % To update the meta ratings, we first compute the rating updates using OpenSkill, resulting in one new rating per player. Since all players from the same team share the same context, this process produces multiple different updates for the same context.  To derive a single updated meta rating for each context, we average the updates of all players sharing that context in the game. Averaging ensures that the meta rating reflects the collective contributions and performances of all players associated with the context. The updates can be written as follows:

    %The meta ratings of all the players belonging to the same context are then updated with OpenSkill. 
    
    %Because multiple players can have the same context in the game, we end up with multiple different rating updates for each context. We choose to average the rating updates. The updates can be written as follows:

    \begin{equation}
        \mu_i^{\text{meta,t+1}} =
    \frac{1}{n} \cdot \sum_{j=1}^n (\mu_j^{\text{meta,t+1}} - \theta_j^{\text{ctx}}),
        \label{eq:update_mu_meta_game}
    \end{equation}

    \begin{equation}
        \sigma_i^{\text{meta,t+1}} = \sqrt{\frac{1}{n} \cdot \sum_{j=1}^n (\sigma_j^{\text{meta,t+1}})^2},
        \label{eq:update_sigma_meta_game}
    \end{equation}

    where $j$ represents a player in the game, and $n$ is the number of players in the game.

    %Contrary to the contextual rating update rule, $\theta_i^{\text{ctx}}$ plays an important role here in taking into account the relative strength of the players in their context. For instance, if players with low contextual ratings from a given context face players with high contextual ratings from another context, the meta rating updates should be different than if all the players had high contextual ratings in their context.
    
\end{itemize}

Note that, in both Equations \ref{eq:openskill_ffa_ctx} and \ref{eq:openskill_ffa_meta}, while we used the FFA OpenSkill setting, we can also use the standard OpenSkill, which bases the updates on the game outcome, if we prefer.

Now that we know how to update the contextual and meta ratings, we need to handle the following edge case. When a player's context changes (e.g., the player goes to another team in another region), their contextual rating might not be representative to the player's skill level in this new context. When that happens, we suggest resetting $\sigma_i^{\text{ctx}}$ to its default value of $25/3$ to represent increased uncertainty in the contextual rating of the player in this new context.

\subsection{Ranking \& Player Comparison}

PandaSkill can be used to rank and compare players. In particular, the probability of player $i$ being better than player $j$ can be estimated using the cumulative distribution function (CDF) $\Phi$:

\begin{equation}
    P(S_i > S_j) = \Phi\left(\frac{\mu_i - \mu_j}{\sqrt{\sigma_i^2 + \sigma_j^2}}\right),
    \label{eq:ffa_probability}
\end{equation}

To rank all the players, instead of using the player comparison probability estimation, which would be too computationally heavy, we can use the lower bound $\theta_i = \mu_i - 3 \cdot \sigma_i$ as a single-value estimate. While this approach is computationally efficient and easy to interpret, it may result in lower $\theta_i$ values for players with fewer games (e.g., new professional players or substitutes) due to higher uncertainty ($\sigma_i$). However, this ensures that the rankings reflect not only performance but also the confidence in the player's skill rating.

% Note that, using this estimate, players with little amount of games (e.g., new professional players, substitute players) will have in average a higher $\sigma_i$ and thus a lower $\theta_i$ than other players.
\section{Experiments}

\label{sec:experiments}

In this section, we describe the experiments we conducted applying PandaSkill to the esports title League of Legends. The general goal is to assess the fairness and relevance of the PScores and skill ratings, both across roles and regions. Because, there is no known ground truth for neither of them, we must rely on proxy metrics and experiments.

\subsection{Data}

\subsubsection{Raw Data}

We used publicly available data from professional League of Legends games obtained through the Leaguepedia API \cite{leaguepedia_api}. The dataset comprises games from all the professional regions worldwide, spanning five years from 2019-09-15 to 2024-09-15. The end date coincides with the end of the last major regional tournaments before the start of Worlds 2024. Table \ref{tab:data} provides a detailed view of the composition of this dataset. We note the large discrepancy between the number of games in inter-region tournaments (Worlds and MSI) compared to regional ones.

\begin{table}[H]
\centering
\caption{Description of the dataset for major competitions.}
\label{tab:data}
\begin{tabular}{lcccc}
\toprule
\textbf{Competition} & \textbf{Region} & \textbf{Editions} & \textbf{Games} & \textbf{Players} \\
\midrule
\textbf{Worlds}  & Global &                     5 &  592   &   353 \\
\textbf{MSI} & Global &     4 &  312  &    166 \\
\textbf{LCK}   & Korea &                      10 & 2,438   &   198 \\
\textbf{LPL}  & China &                       12 & 3,643   &   283 \\
\textbf{LEC}   & Europe          &            15 & 1,315   &   149 \\
\textbf{LCS}  & North America      &                 13 & 1,371   &   182 \\
\textbf{PCS}  & Asia-Pacific     &                  10 & 1,295   &   235 \\
\textbf{VCS}    & Vietnam        &             12 & 1,489   &   158 \\
\textbf{CBLOL}  & Brazil         &            10 & 1,153   &   196 \\
\textbf{LLA}     & Latin America       &             12 &  890   &   175 \\
\multicolumn{5}{c}{\dots} \\
\midrule
\textbf{Total}              & & \textbf{392} & \textbf{37,388} & \textbf{4,927} \\

\bottomrule
\end{tabular}
\end{table}

\begin{table*}[!ht]
\centering
\caption{Description of player statistics used as input features to the performance score models.}
\label{tab:feature-descriptions}
\begin{tabular}{ll}
\toprule
\textbf{Feature} & \textbf{Description} \\
\midrule
\textbf{Kill-life-assist ratio (KLA)} & Ratio of enemy player kills compared to deaths. Computed as $(kills + assists) / (deaths + 1)$. \\
\textbf{Gold per minute} & Rate at which the player obtained gold in the game. \\
\textbf{Experience per minute} & Rate at which the player gained experience in the game. \\
\textbf{Creep score per minute} & Rate at which the player slays Minions or neutral monsters, procuring gold. \\
\textbf{Wards placed per minute} & Rate at which the player placed wards to provide vision for the team. \\
\textbf{Damage dealt total kills ratio} & Amount of damage dealt to players, normalized by the total number of kills in the game. \\ 
\textbf{Damage dealt per gold total kills ratio} & Player damage dealing efficiency, computed as \textit{Damage dealt total kills ratio} further normalized by gold obtained. \\ 
\textbf{Damage taken total kills ratio} & Amount of damage taken from players, normalized by the total number of kills in the game. \\ 
\textbf{Damage taken per gold total kills ratio} & Player damage tanking efficiency, computed as \textit{Damage taken total kills ratio} further normalized by gold obtained. \\ 
\textbf{Largest multi kill} & Maximum number of player kills in a short time window. \\
\textbf{Largest killing spree total kills ratio} & Maximum number of successive player kills without dying, normalized by the total number of kills in the game. \\ 
\textbf{Worthless death ratio} & Ratio of player deaths that did not benefit the team in any way. \\ 
\textbf{Free kill ratio} & Ratio of player kills that resulted in a worthless death for the enemy. \\ 
\textbf{Objective contest winrate} & Share of neutral objectives contested and won by the player. \\ 
\textbf{Objective contest loserate} & Share of neutral objectives contested but lost by the player. \\ 
\bottomrule
\end{tabular}
\end{table*}

For each game, we have the raw end-game statistics of the players (e.g., amount of gold, number of kills), events (e.g., player killing another player), game-related data (e.g., duration), and metadata (e.g., competition the game was in). Each player in a game has been assigned a region based on their participation in the latest highest-tier regional tournaments.

\subsubsection{Feature Extraction}

From the raw data, we extracted features that both reflect metrics widely used by players and facilitate modeling. Table \ref{tab:feature-descriptions} describes all the features we used. 

Most of the data points are normalized using reference values specific to each game. The two most common normalizations are the game-length normalization and total-kills normalization. Game-length normalization accounts for different game durations (e.g., having 10 kills in 15 minutes is not the same as in 50 minutes). Total-kills normalization accounts for different pacing in the game, as some games can have much more fighting than others.

%Most of the data points have been normalized relative to another game value. The two most common normalizations are the game-length normalization and total-kills normalization. Game-length normalization accounts for different game durations (e.g., having 10 kills in 15 minutes is not the same as in 50 minutes). Total-kills normalization accounts for different pacing in the game, as some games can have much more fighting than others.

Particular attention has been given to using features or data points that are strongly linked solely to the player's performance. Because of that, we have excluded all team-related statistics (e.g, total number of kills by the team), and decided against using team-normalization techniques (e.g., share of number of kills within the team).

Some features described in Table \ref{tab:feature-descriptions} deserve more attention: 

\begin{itemize}
    \item \textbf{Kill-life-assist ratio (KLA)}: This formula has been preferred over the more traditional Kill-death-assist ratio (KDA) for numerical stability. Indeed, both formulas $KDA = (kills + assists) / deaths$ or $KDA = (kills + assists) / max(deaths, 1)$ are either instable when $deaths = 0$ or cannot differentiate between 0 and 1 death.
    \item \textbf{Worthless death ratio and Free kill ratio}: A player's death is considered worthless if, within a 1-minute window, the player has not been involved in an enemy kill or their team has not secured an objective (e.g., a Drake or a Tower). The latter describes situations where the player's death gave the opportunity to their team to accomplish something else important on the map. Note that, while it shares the same core idea as the worthless death described by Maymin \cite{maymin2021smart}, we preferred a less complex implementation as it is not the focus of the paper.
    %\item \textbf{Objective contest winrate and loserate}: We compute these features from the neutral monster kill events. If players from both teams are found among the killer or assists of the event, then the event is contested. The contest win goes to the team of the killer. To compute the rates, we normalize the number of contest wins and losses by the total number of contestable objectives in the game.
    \item \textbf{Objective contest winrate and loserate}: These features are computed based on neutral monster kill events. An event is considered contested if players from both teams are present among the killer or assists of the event, with the contest win assigned to the team of the killer. The rates are calculated by normalizing the number of contest wins and losses by the total number of contestable objectives in the game. It is important to note that, due to the lack of event location data in the dataset, fights occurring near a neutral objective are not categorized as contests for that objective unless explicitly registered as part of the neutral monster kill event.
\end{itemize}

\subsection{Computing and Evaluating the Players' Game Performance}

First, we consider the computation of the player's performance scores.

\subsubsection{Implementation Details}  As discussed in Section \ref{sec:methods_perf}, we implemented our proposed approach using multiple XGBoost models, one per role. Specifically, they were trained with 2,000 boosted rounds and a learning rate of 0.01. 

To improve the interpretability of the model, we enforce monotonicity constraints for all the features. These constraints force each feature to impact the output probability in only one direction. For instance, an increase in the KLA should only increase the win probability, not decrease it. This prevents the model from finding hidden patterns in the combination of features that could lead to unexpected uses of certain features. In practice, of all the features we used, only the Worthless death ratio and Objective contest loserate are forced to decrease the output probability with higher values.

A 5-fold cross-validation was used to compute the performance of the players and evaluate the models. Before being input to the models, all the features were standardized to zero mean and unit variance. 

% Our models have an average accuracy of 90.74\% (Standard Deviation, SD = 0.60) in predicting the game outcome with the monotonicity constraints enforced, and 91.79\% (SD = 0.56) without them. These results are comparable to role-based SVC models (91.00\% accuracy, SD = 0.46), as in the PlayeRank framework \cite{pappalardo2019playerank}, or to Random Forest models (91.30\% accuracy, SD = 0.58) as in the PI framework \cite{demediuk2021performance}. These high accuracy values are expected since we are predicting the outcome of the game based on the end-game statistics, which contain extensive information about what happened during the game.

\subsubsection{Experimental Results} Evaluating the performance scores is difficult because of the lack of known ground truth. In these experiments, three aspects of the models are looked at: their predictive capability, the calibration of the predicted probabilities, and the interpretability of the outputs.

First, regarding the predictive capability, our XGBoost models achieve an average accuracy of 90.74\% (Standard Deviation, SD = 0.60), comparable to role-based SVC models (91.00\% accuracy, SD = 0.46) from the PlayeRank framework \cite{pappalardo2019playerank}, and Random Forest models (91.30\% accuracy, SD = 0.58) from the PI framework \cite{demediuk2021performance}. These high accuracy values are expected, as we are predicting game outcomes based on end-game statistics, which contain extensive information about in-game events. The slightly lower accuracy of our models can be attributed to the enforcement of monotonicity constraints. Without these constraints, our models achieve a higher accuracy of 91.79\% (SD = 0.56). However, this improved performance comes at the cost of using unintended feature interactions, which undermine the explainability of the performance scores (e.g., a higher creep score per minute paradoxically decreasing the predicted winning probability).

\begin{figure}[htbp]
    \centering
    \includegraphics[width=0.45 \textwidth]{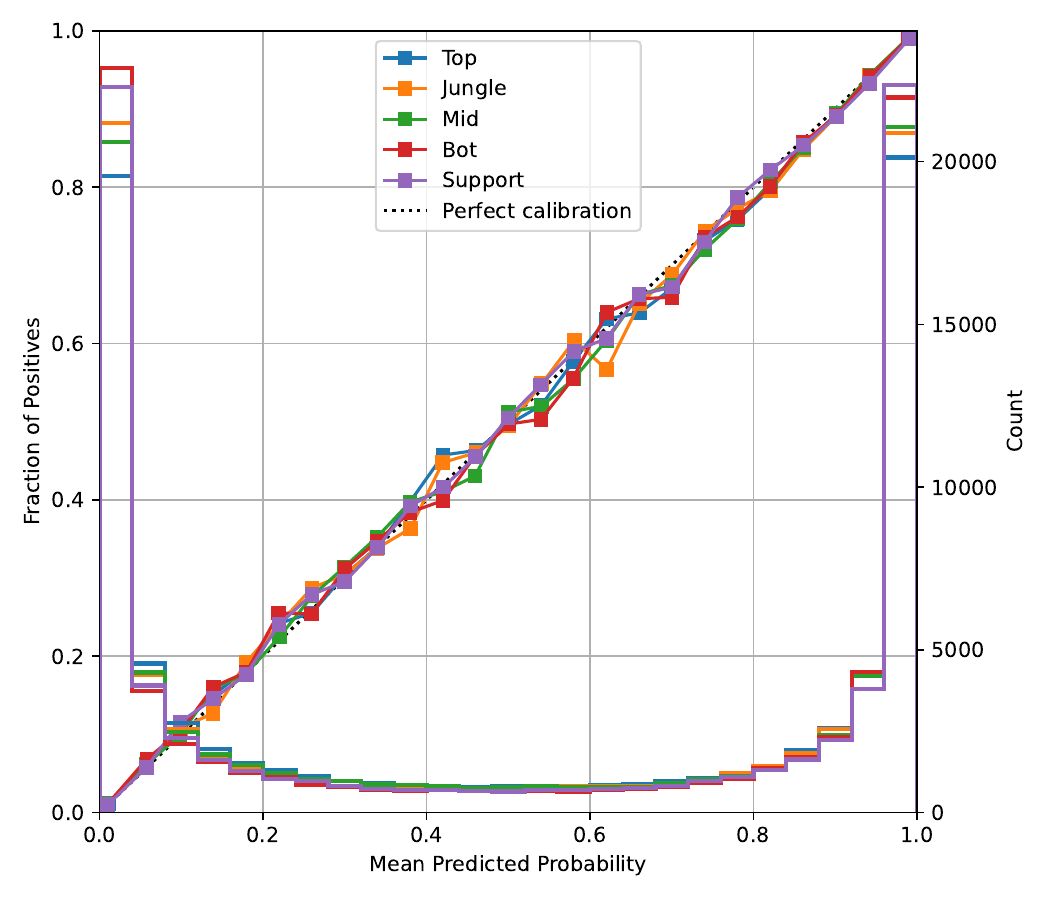}
    \caption{Calibration plots with histograms of predicted probabilities for each role.}
    \label{fig:calibration_plots}
\end{figure}

As mentioned in section \ref{sec:methods_perf}, the good calibration of the model is very important, as the player performance is directly related to the trustworthiness of the probabilities. We used the Expected Calibration Errror (ECE) to measure quality of the estimated probabilities \cite{silva2023classifier}. With a mean ECE ( of 0.93\% (SD = 0.03), the model is fairly well calibrated. In comparison, the PlayeRank approach has an average ECE of 1.28\% (SD = 0.13) and the PI approach an average ECE of 2.28\% (SD = 0.09). Figure \ref{fig:calibration_plots} displays the calibration plots of our models for the different roles, alongside the histogram of the probability values. The U-shape of the probability distribution, combined with its variability across different roles, shows the need for the percentile transformation in the calculation of PScore to achieve uniformly distributed game performances across all roles.

\begin{figure*}[htbp]
    \centering
    \includegraphics[width=0.9\textwidth]{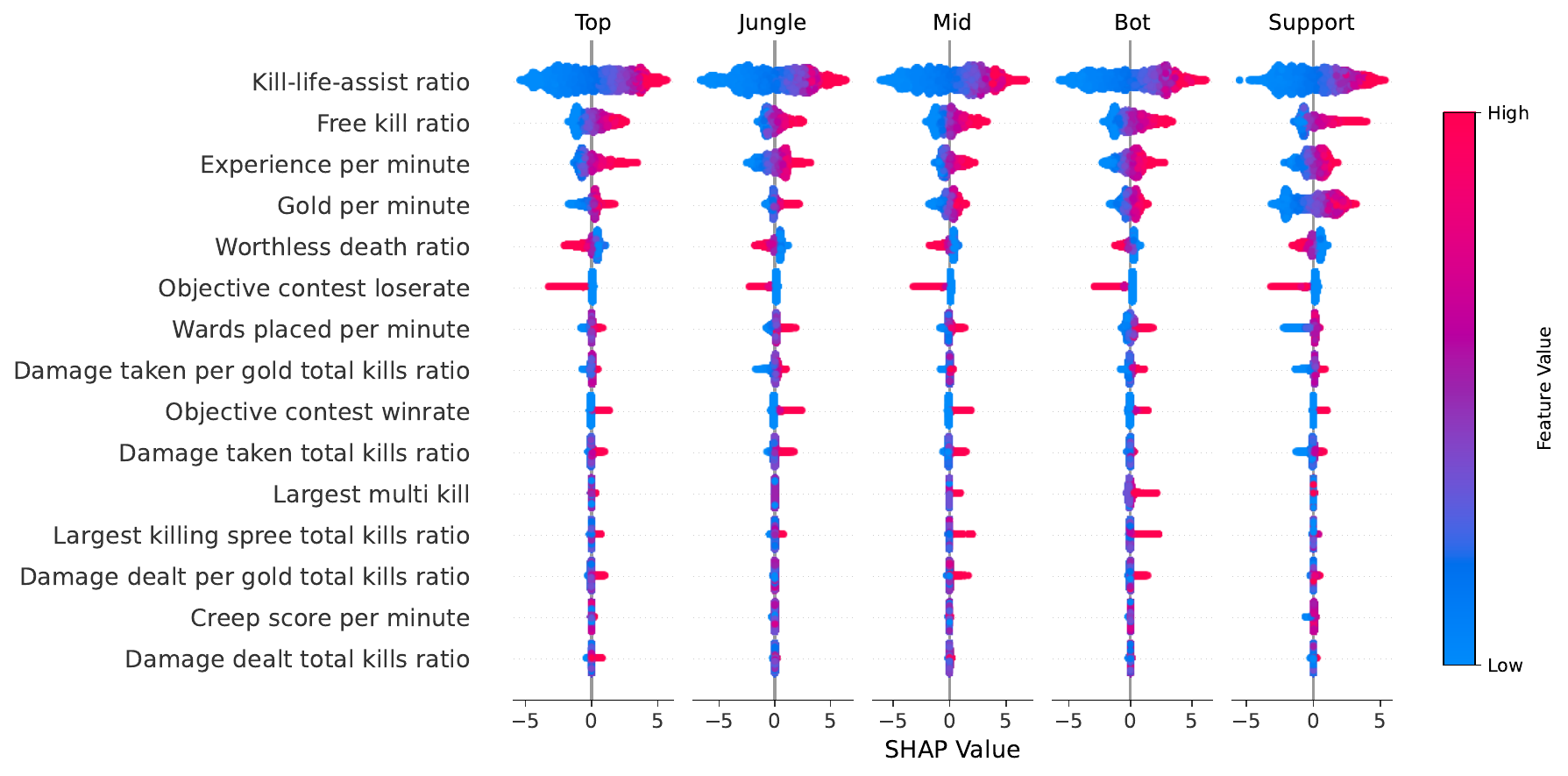}
    \caption{Relative importance of features for each role-based XGBoost model using SHAP values. The features are ordered in descending order based on their average importance across roles.}
    \label{fig:feature_importance}
\end{figure*}

Finally, we examine the importance of the different features used to compute the performance scores. To quantify each feature's contribution to the predictions, we use SHAP (SHapley Additive exPlanations) values \cite{lundberg2017unified}. SHAP is a model-agnostic explainability method that measures the average change in a model's predictions when a given feature is included or excluded. SHAP values can explain individual predictions, as shown in Table \ref{table:game_analysis}, or represent general behavior through their distribution over games and roles, as illustrated in Figure \ref{fig:feature_importance}. The KLA emerges as the most important feature across all roles. Other significant features include the Free kill ratio, Experience per minute, Gold per minute, Worthless death ratio, and Objective contest loserate. This highlights not only the importance of well-known metrics like Gold per minute but also the relevance of new features we introduced, such as the Free kill ratio. Besides, Figure \ref{fig:feature_importance} shows the impact of the monotonicity constraints that have been enforced in the XGBoost models, as all the features are pushing the winning probabilities in the expected directions.

%One benefit of using an XGBoost implementation is the interpretability of the predictions. To quantify the contribution of each feature to the prediction, we employ SHAP (SHapley Additive exPlanations) values \cite{lundberg2017unified}. By computing the average change in the model's predictions when including or excluding a given feature, SHAP values assign an importance score to each feature. Figure \ref{fig:feature_importance} illustrates the distribution of SHAP values for different roles, ordered by the global importance of each feature. The KLA is indisputably the most important feature for all roles. After that, the importance of the features varies depending on the role. For instance, the Gold per minute feature is% really important for the Support role. This means that if a Support player has a high Gold per minute value, then the player has most likely won the game. This is explained by the fact that, since they do not farm gold from Minions, most of the gold for Support players comes from objectives and player kills, which is a strong indicator of the game outcome.

To conclude, our approach implemented with XGBoost has a good predictive capability while being calibrated and maintaining a good interpretability. The next sets of experiments will further validate the approach by showing the usefulness of the PScore values in the creation of accurate skill ratings.

\subsection{Computing and Evaluating the Player Skill Ratings}

\subsubsection{Experimental Models}

In this section, we are interested in how well our OpenSkill-based framework estimates the player's skill ratings. To measure the impact of the different modules we proposed, we consider the following model variations:

%In this section, we are interested in how well the player's skill ratings are estimated using the OpenSkill-based framework we proposed. Five rating model variations are considered:
\begin{itemize}
    \item \textbf{OpenSkill}: The skill rating updates only depend on the team outcome of the game and not the player's in-game performance.
    \item \textbf{FFA\_OpenSkill}: The skill rating updates use the FFA setting described in Section \ref{sec:ffa}.
    \item \textbf{Meta\_OpenSkill}: The skill rating is defined as the combination of both contextual and meta ratings as described in Section \ref{sec:meta_rating}.
    \item \textbf{Meta\_FFA\_OpenSkill}: The skill rating is both defined as the combination of both contextual and meta rating, and use the FFA setting in the update.
    \item \textbf{Meta\_FFA\_TrueSkill}: Similar to the Meta\_FFA\_OpenSkill variation, but with TrueSkill instead of OpenSkill. We specifically used the non official implementation of Lee, as the official one is not available \cite{lee2018trueskill,Joshy2024}.
\end{itemize}

In addition, as it is used in the PlayeRank framework \cite{pappalardo2019playerank}, we compare these rating models to EWMA, which we implemented with a smoothing factor of 0.05 to put more emphasis on the player's performance history. 

We apply these rating models to the PScore, PlayeRank, and PI performance models.

\subsubsection{Pre-match Game Outcome Forecasting Experiment}

First, we evaluate the models by their ability to forecast the outcome of the game from the players' skill ratings before the game. The intuition is that more accurate skill ratings should be better predictors of the outcome of the game. 

To account for the gradual changes in the rating distribution and potential meta shifts in the video game, we used a rolling window of one month. We trained a logistic regression model on one year of data and tested it on the subsequent month. Then, we shifted our window by one month and repeated the training-testing evaluation.

\begin{figure}[htbp]
    \centering
    \includegraphics[width=0.45\textwidth]{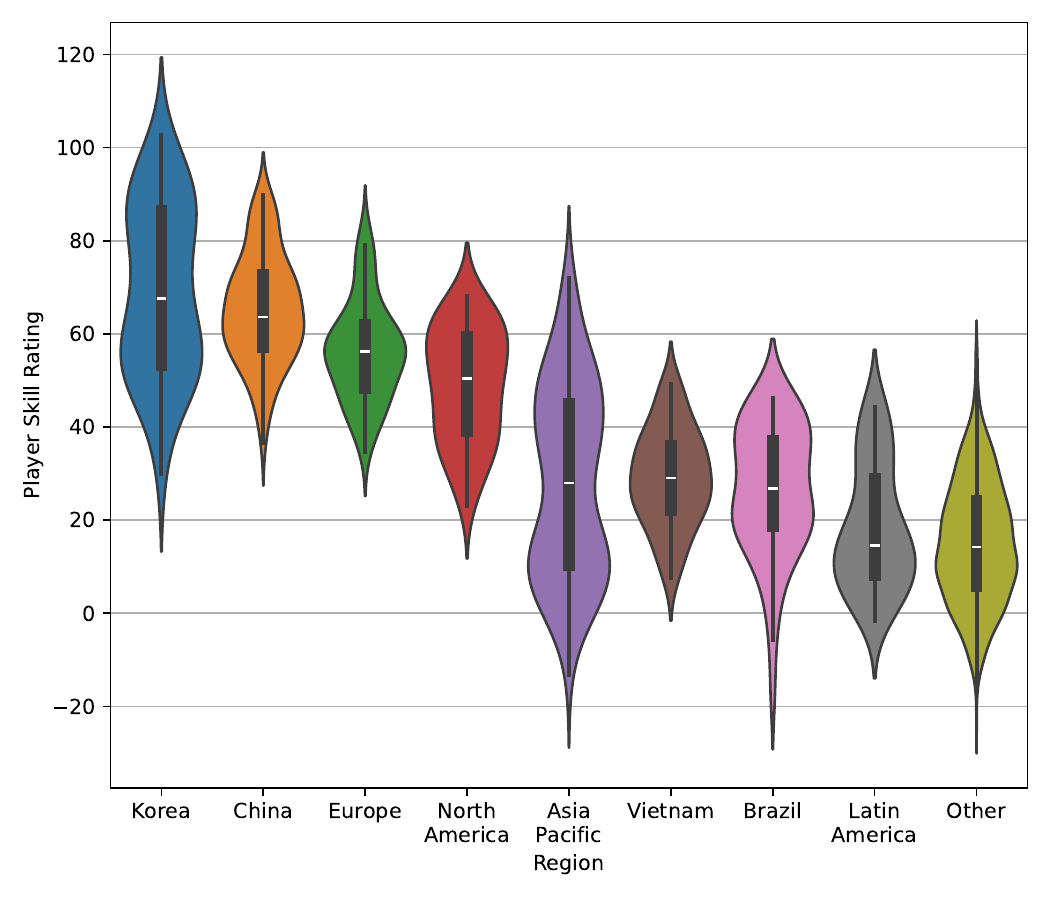}
    \caption{Distribution of players' skill ratings per region using the Meta\_FFA\_OpenSkill rating model with PScore.}
    \label{fig:region_ratings}
\end{figure}

Table \ref{tab:outcome_from_ratings} shows the performance of the models in terms of accuracy and ECE. Results are detailed for both intra-region and inter-region matches. We can observe the following:
\begin{itemize}
    \item All the OpenSkill (or TrueSkill) variations have a higher accuracy than their EWMA counterparts. This shows the benefits of having Bayesian ratings with updates depending on the skill ratings of the opponents.
    \item All the models implementing the Meta OpenSkill variation see a large increase in accuracy in inter-region matches, without sacrificing much in intra-region accuracy. Inter-region accuracy becomes higher than intra-region accuracy, which makes sense given how spread apart the skill of different regions can be (and thus easier to predict), as shown by Figure \ref{fig:region_ratings}.
    \item The FFA OpenSkill variation observes a small decrease in accuracy compared to the non-FFA OpenSkill model. This is not surprising, as for this experiment, only a good overall team rating is needed.
    \item The improvements of the proposed rating model transfer well to other game performance models, as shown by the PI + Meta\_FFA\_OpenSkill and PlayeRank + Meta\_FFA\_OpenSkill model combinations.
    \item Compared to OpenSkill, TrueSkill obtains comparable accuracy, but significantly worse calibration.
\end{itemize}

\begin{table*}[!ht]
\centering
\caption{Accuracy and ECE of a logistic regression model forecasting the game outcome from the players' skill ratings. Results are described for all the games, or for intra-region or inter-region games only.}
\label{tab:outcome_from_ratings}
\begin{tabular}{lcccccc}
\toprule
\multirow{2}{*}{\textbf{Model}} & \multicolumn{3}{c}{\textbf{Accuracy (\%) $\uparrow$}} & \multicolumn{3}{c}{\textbf{ECE (\%) $\downarrow$}} \\ 
\cmidrule(lr){2-7} & All & Intra & Inter & All & Intra & Inter \\

\toprule
\textbf{PScore + EWMA} & 63.06 & 63.33 & 52.34 & 1.43 & 1.57 & 5.51 \\ 
\textbf{PScore + OpenSkill} &  \textbf{65.56} & \textbf{65.51} &  67.39 & 1.53 & 1.39 & 7.59\\ 
\textbf{PScore + Meta\_OpenSkill} & 65.12 & 65.01 & 69.23 & 0.89 & 0.88 & 5.51\\ 
\textbf{PScore + FFA\_OpenSkill} & 64.79 & 64.86 & 61.71 & \textbf{0.77} & \textbf{0.79} & 4.02\\ 
\textbf{PScore + Meta\_FFA\_OpenSkill} & 64.98 & 64.86 & 70.07 & 1.01 & 0.97 & \textbf{3.27}\\ 
\textbf{PScore + Meta\_FFA\_TrueSkill} & 65.15 & 65.05 & 68.90 & 2.85 & 2.85 & 4.85\\ 
\midrule
\textbf{PI + EWMA} & 63.53 & 63.75 & 54.68 & 1.88 & 1.99 & 4.61 \\ 
\textbf{PI + Meta\_FFA\_OpenSkill} & 65.05 & 64.92& \textbf{70.23} & 1.00& 1.05 & 4.15\\ 
\midrule
\textbf{PlayeRank + EWMA} & 62.25 & 62.80 & 51.67 & 2.03 & 2.22 & 6.16\\  
\textbf{PlayeRank + Meta\_FFA\_OpenSkill} & 64.94 & 64.84 & 69.23 & 1.01 & 0.97 & 5.40 \\ 
\bottomrule
\end{tabular}
\end{table*}

\subsubsection{Skill Ratings Distribution per Role} 

%One important aspect of PandaSkill, from the evaluation of the player's in-game performance of players, to the computation of their skill ratings, is that all roles are treated identically. We use the Wasserstein distance between the rating distributions of two roles, averaged over every pair of role, to measure this fairness of treatment \cite{panaretos2019statistical}.

%First, all PlayeRank-based approaches perform poorly, with the best one being PlayeRank + EWMA, which has an average of 2.44 (SD = 1.32). On the other hand, all PScore-based models have values between 0.09 (SD = 0.01) for the Meta\_FFA\_TrueSkill variation and 0.44 (SD = 0.13) for the Meta\_FFA\_OpenSkill variation. As for PI-based models, they perform between 0.35 (SD = 0.12) and 0.66 (SD = 0.21) for the EWMA and Meta\_FFA\_OpenSkill models respectively, which is slightly worse than PScore but still much better than PlayeRank.

One important aspect of PandaSkill, from the evaluation of players' in-game performance to the computation of their skill ratings, is that all roles are treated identically. To measure this fairness of treatment, we use the Wasserstein distance between the rating distributions of two roles, averaged over every pair of roles. The Wasserstein distance measures the minimal 'effort' required to transform one probability distribution into another, effectively quantifying differences between distributions \cite{panaretos2019statistical}.

Results, grouped by performance score model, are as follows:
\begin{itemize}
    \item {\textbf{PlayeRank}}: Performs poorly, with the best one averaging a Wasserstein distance of 2.44 (SD = 1.32).
    \item {\textbf{PI}}: Shows good performance, with distances ranging from 0.35 (SD = 0.12) to 0.66 (SD = 0.21).
    \item {\textbf{PScore}}: Achieves the best performance, with distances between 0.09 (SD = 0.01) and 0.44 (SD = 0.13).
\end{itemize}

First, both PScore-based and PI-based models significantly outperform PlayeRank-based approaches. This can be attributed to the use of percentile transformations. While PI applies this transformation to the input features, PandaSkill applies it to the output performance scores. In both cases, roles are treated more fairly as the scale of either the model inputs or outputs becomes independent of the role. However, from the lower average distance of PScore, we can conclude that doing the percentile transformation at the end, as in PandaSkill, is more effective. 

\subsection{Evaluating the Player Ranking}

Table \ref{tab:ranking} shows the top 50 players from all the regions. We can see it is dominated by Korea and China, which aligns with most of the inter-region results from the past few years (e.g., MSI or Worlds). Only one Western team, G2 Esports, made it into the top 50 ranking.

To perform a quantitative evaluation of the rankings and the underlying skill ratings, as done in the evaluation of PlayeRank \cite{pappalardo2019playerank}, we conducted an evaluation by human experts.

%regex for global ranking csv
%(\d+),.*,(.*),(.+),(.*),.*,(.*),.*,.*,(.*),.*,.*
%$1 & $2 & $3 & $4 & $5 & $6 \\\\

%& (\d+\.\d\d)\d+ \\
%& $1 \\

\begin{table*}[ht]
    \centering
    \begin{tabular}{c c}
        % First table
    \begin{threeparttable}
    \begin{tabular}{c|ccccc}
      \toprule
      \textbf{Rank} & \textbf{Player} & \textbf{Team} & \textbf{Region} & \textbf{Role} & \textbf{Rating} \\
      \midrule
        1 & Chovy & Gen.G & Korea & Mid & 102.81 \\
        2 & Peyz & Gen.G & Korea & Bot  & 101.31 \\
        3 & Canyon & Gen.G & Korea & Jungle & 97.96 \\
        4 & Kiin & Gen.G & Korea & Top & 96.10 \\
        5 & Viper & HLE\tnote{1} & Korea & Bot  & 94.37 \\
        6 & Delight & HLE\tnote{1} & Korea & Support & 91.93 \\
        7 & Peanut & HLE\tnote{1} & Korea & Jungle & 91.92 \\
        8 & Lehends & Gen.G & Korea & Support & 91.92 \\
        9 & Zeka & HLE\tnote{1} & Korea & Mid & 91.31 \\
        10 & ON & BLG\tnote{2} & China & Support & 89.85 \\
        11 & Aiming & Dplus KIA & Korea & Bot  & 89.73 \\
        12 & Oner & T1 & Korea & Jungle & 89.08 \\
        13 & Elk & BLG\tnote{2} & China & Bot  & 89.06 \\
        14 & knight & BLG\tnote{2} & China & Mid & 88.64 \\
        15 & Keria & T1 & Korea & Support & 88.11 \\
        16 & Gumayusi & T1 & Korea & Bot  & 87.93 \\
        17 & Tian & Top Esports & China & Jungle & 87.90 \\
        18 & Lucid & Dplus KIA & Korea & Jungle & 86.56 \\
        19 & Doran & HLE\tnote{1} & Korea & Top & 86.23 \\
        20 & Bin & BLG\tnote{2} & China & Top & 86.18 \\
        21 & Meiko & Top Esports & China & Support & 86.12 \\
        22 & 369 & Top Esports & China & Top & 86.08 \\
        23 & Kanavi & JD Gaming & China & Jungle & 86.02 \\
        24 & Zeus & T1 & Korea & Top & 85.73 \\
        25 & Ruler & JD Gaming & China & Bot  & 85.26 \\
      \bottomrule
    \end{tabular}
    \begin{tablenotes}
    \item[1] Hanwha Life Esports.
    \item[2] Bilibili Gaming.
    \end{tablenotes}
    \end{threeparttable}
        &
        % Second table
    \begin{threeparttable}
    \begin{tabular}{c|ccccc}
      \toprule
      \textbf{Rank} & \textbf{Player} & \textbf{Team} & \textbf{Region} & \textbf{Role} & \textbf{Rating} \\
      \midrule
        26 & Missing & JD Gaming\phantom{\tnote{1}} & China & Support & 84.24 \\
        27 & XUN & BLG\tnote{2} & China & Jungle & 83.89 \\
        28 & Faker & T1\phantom{\tnote{1}} & Korea & Mid & 83.49 \\
        29 & Creme & Top Esports\phantom{\tnote{1}} & China & Mid & 82.61 \\
        30 & Hans sama & G2 Esports\phantom{\tnote{1}} & Europe & Bot  & 82.36 \\
        31 & JackeyLove & Top Esports\phantom{\tnote{1}} & China & Bot  & 82.32 \\
        32 & Bdd & KT Rolster\phantom{\tnote{1}} & Korea & Mid & 81.84 \\
        33 & ShowMaker & Dplus KIA\phantom{\tnote{2}} & Korea & Mid & 81.80 \\
        34 & Flandre & JD Gaming\phantom{\tnote{2}} & China & Top & 80.36 \\
        35 & Yagao & JD Gaming\phantom{\tnote{2}} & China & Mid & 80.18 \\
        36 & BrokenBlade & G2 Esports & Europe & Top & 79.11 \\
        37 & Hang & LNG Esports & China & Support & 78.94 \\
        38 & Scout & LNG Esports & China & Mid & 78.94 \\
        39 & Pyosik & KT Rolster & Korea & Jungle & 78.86 \\
        40 & GALA & LNG Esports & China & Bot  & 78.74 \\
        41 & Caps & G2 Esports & Europe & Mid & 78.20 \\
        42 & BeryL & KT Rolster & Korea & Support & 78.13 \\
        43 & Deft & KT Rolster & Korea & Bot  & 77.74 \\
        44 & Kellin & Dplus KIA & Korea & Support & 77.61 \\
        45 & Tarzan & Weibo Gaming & China & Jungle & 76.91 \\
        46 & Zika & LNG Esports & China & Top & 76.83 \\
        47 & Yike & G2 Esports & Europe & Jungle & 76.73 \\
        48 & Light & Weibo Gaming & China & Bot  & 76.71 \\
        49 & Mikyx & G2 Esports & Europe & Support & 76.13 \\
        50 & Weiwei & LNG Esports & China & Jungle & 75.94 \\
      \bottomrule
    \end{tabular}
    \end{threeparttable}
    \end{tabular}
    \caption{Ranking for the first 50 players using the lower bound of their skill rating.}
    \label{tab:ranking}
\end{table*}

\subsubsection{Experimental Setting}

We created several surveys, each containing 300 randomly selected pairs of players, and asked experts to choose the better player from each pair. Five surveys were created: four focused on the major regions (Korea, China, Europe, and North America), and one global survey with players from all regions. The experts, recruited from PandaScore’s odds traders, were selected based on their expertise in the different LoL regions. Since not all traders were equally familiar with every region, the regional surveys were only completed by those with sufficient knowledge. Table \ref{tab:surveys} details the composition of these surveys. 

To analyze the results, we investigated two different types of concordance between the model and the experts: 
\begin{itemize}
    \item \textbf{Majority concordance}: The fraction of pairs for which a given model agrees with the majority of the experts.
    \item \textbf{Unanimity concordance}: The fraction of pairs for which a given model agrees with all the experts.
\end{itemize}

In cases where no majority could be achieved among the experts (e.g., when the number of experts was even, and exactly half of them preferred one player), we chose to drop those comparisons (represented by the \textit{No majority} column in Table \ref{tab:surveys}).

\begin{table}[H]
\centering
\caption{Description of player comparison surveys with number of experts, number of unique players, ratio of unanimous answers, ratio of no majority answers.}
\label{tab:surveys}
\begin{tabular}{lcccc}
\toprule
\textbf{Region} & \textbf{Experts} & \textbf{Players} & \textbf{Unanimity} & \textbf{No majority} \\
\midrule
Global & 5 & 477 & 0.51 & -\\
Korea & 4 & 54 & 0.70 & 0.06\\
China & 3 & 96 & 0.59 & - \\
Europe & 4 & 52 & 0.62 & 0.17\\
North America & 3 & 40 & 0.59 & -\\
\midrule
\textbf{Total} & 5 & 629 & 0.60 & 0.05 \\
\bottomrule
\end{tabular}
\end{table}

\subsubsection{Results}

 The PScore + Meta\_FFA\_OpenSkill model achieves the second- highest majority concordance (average of 80.63\%, SD = 6.52), just behind PI + Meta\_FFA\_OpenSkill, which has 80.70\% (SD = 5.41). However, it achieves the highest average unanimity concordance with an average of 88.98 \% (SD = 5.99). 

\begin{figure*}[htbp]
    \centering
    \includegraphics[width=\textwidth]{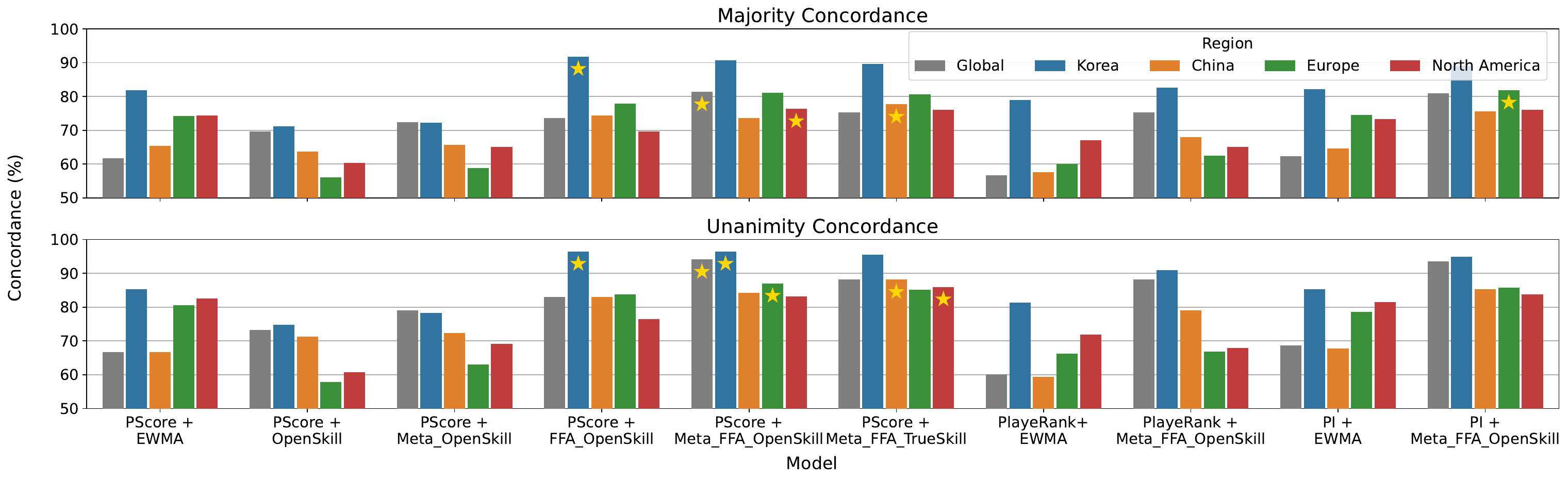}
    \caption{Majority and unanimity concordance between models and experts per region.}
    \label{fig:ranking_eval_region}
\end{figure*}

Figure \ref{fig:ranking_eval_region} details the results per model and per region:
\begin{itemize}
    \item EWMA is outperformed by the Meta\_FFA\_OpenSkill rating model for almost all underlying performance models and regions.
    \item Using the FFA OpenSkill setting shows a significant improvement in the concordance with the experts, demonstrating the usefulness of incorporating the player's in-game performance in the updates of their skill ratings.
    \item Using the dual-rating system with a contextual rating and a meta rating, alongside the FFA setting, significantly increases the concordance for the Global survey. This highlights the issue of isolated rating pools coming from the low number of inter-region games. PandaSkill provides a working solution to this issue.
    \item TrueSkill is comparable to OpenSkill, being only notably worse for the Global survey. This indicates that the dual-rating system we propose works better with the OpenSkill framework.
    \item PI-based player performance ultimately offers very similar skill ratings compared to PScore. This shows that PScore is a valid alternative to PI, with the advantage of being model-agnostic by design.
    \item Even when improved by the Meta\_FFA\_OpenSkill rating model, PlayeRank still fails to offer convincing skill ratings compared to other approaches. This underscores the need for a good underlying player performance model.
    \item Surprisingly, the concordance between experts and the models is very high for the Korea region. This can be explained by the wider range of skill ratings that is within the region compared to the others, as shown by Figure \ref{fig:region_ratings}.
\end{itemize}

\subsection{Example of the Framework Applied onto a Game}

To demonstrate PandaSkill's practical usage, we propose to look at how it behaves in a given game. We chose game 2 of the match between Hanwha Life Esports (HLE) and T1 in the lower-bracket final of the LCK Summer 2024. It was a high-stake game, between two of the strongest teams in the world (see Table \ref{tab:ranking}). The game was very close and ended with the victory for T1.

As shown in Table \ref{table:game_analysis}, despite losing the game, the skill ratings of multiple players from HLE (Viper, Peanut, and Zeka) increased after the game. This was due to the strong performances of these players compared to most of T1's players. Conversely, the skill ratings of some T1 players decreased (Oner, Zeus). 

Faker was the Most Valuable Player (MVP) of his team, having the highest PScore, with a difference of 21.93 compared to Zeka, the player in the same role for HLE (Mid). Looking at the feature contributions to Faker's performance, around half of his PScore can be attributed to his exceptional KLA (7.0). Other outstanding features are his Gold per minute, Largest multi kill, and Damage taken total kills ratio.

\section{Conclusion}

In this paper, we presented PandaSkill, a framework to compute the performance of players in a game (PScore) and their skill ratings. Compared to other approaches in the literature, PandaSkill is unique in that it fairly compares players from different roles, is model-agnostic, and addresses the isolated rating pools issue common in esports. We applied PandaSkill to five years of professional League of Legends, demonstrating its usefulness.

We have open-sourced PandaSkill so that it can be reused by the community, driving future research. Here are some directions we think could be worth studying. Because our approach is model-agnostic, more complex models such as neural networks could be envisaged \cite{delalleau2012beyond}. The input features to the model are as important, if not more so, than the model itself. The usefulness of more complex features linked to the impact of player actions could be worth exploring \cite{maymin2021smart, xenopoulos2020valuing}. Last but not least, applying PandaSkill to other esports titles such as Counter-Strike or Dota 2 should yield interesting results. While similar to League of Legends in essence, both games have specific aspects that would need to be addressed. For instance, the conception of roles is quite different, potentially being more fluid and less defined than in LoL \cite{varga2024esports, demediuk2021performance}. Moreover, the competitive scenes are fragmented differently. Instead of being fragmented by region, they are fragmented by tiers of competition.

\label{sec:conclusion}

\bibliography{references}
\bibliographystyle{IEEEtran}

\vfill

\end{document}